\documentclass{article}

\usepackage[preprint]{neurips_2026}
\usepackage{xcolor}

\usepackage[hyphens]{url}
\usepackage{hyperref}
\hypersetup{
    colorlinks,
    citecolor=blue,
    filecolor=blue,
    linkcolor=blue,
    urlcolor=blue,
}

\usepackage{enumitem}

\usepackage{makecell}
\usepackage{subcaption}

\usepackage{graphicx}

\usepackage{amsmath}
\usepackage{amssymb}
\usepackage{amsfonts}
\usepackage{amsthm}
\usepackage{bm}
\usepackage{dsfont}

\definecolor{matplotlib_blue}{HTML}{1F77B4}
\definecolor{matplotlib_orange}{HTML}{FF7F0E}
\definecolor{matplotlib_green}{HTML}{2CA02C}
\definecolor{matplotlib_red}{HTML}{D62728}

\newcommand{\operation}[1]{\operatorname{#1}}

\title{Normal Guidance is what Attention Needs}

\author{%
    Ethan Harvey\thanks{Equal contribution} \\
    Department of Computer Science\\
    Tufts University\\
    \texttt{ethan.harvey@tufts.edu} \\
    \And
    Dennis Johan Loevlie\footnotemark[1] \\
    Department of Computer Science\\
    Tufts University\\
    \texttt{dennis.loevlie@tufts.edu} \\
    \And
    Michael C. Hughes \\
    Department of Computer Science\\
    Tufts University\\
    \texttt{michael.hughes@tufts.edu} \\
}

\begin{document}

\maketitle

\begin{abstract}    
We consider training classifiers for 3D medical images using only one binary label for the entire volume rather than a label for each 2D slice. In such weakly supervised settings, can we learn accurate classifiers for slice-level predictions? Attention-based multiple instance learning (MIL) can produce an attention score for every slice. Yet recent work demonstrates that a simple center-focused baseline that ignores image content can outperform attention-based and transformer-based MIL at slice-level classification of 3D brain scans. We show this baseline also outperforms existing MIL at slice-level classification of thoracic and abdominal CT scans. Motivated by this baseline, we propose Normal Guidance, a regularization technique that encourages the learned attention distribution to follow a bell-shaped curve. Across three medical imaging datasets totaling over 4 million 2D slices, we show our Normal Guidance enables attention-based and transformer-based MIL methods to deliver significantly better slice-level localization than the state-of-the-art while remaining competitive at whole-scan classification.
\end{abstract}



%

\let\thefootnote\relax\footnotetext{Code: \href{https://github.com/tufts-ml/normal-guidance}{\texttt{github.com/tufts-ml/normal-guidance}}}

\section{Introduction}
\label{sec:introduction}

Deep learning has become a go-to approach for prediction tasks involving 3D medical images~\citep{wang2017chestx,saab2019doubly,wu2021combining,castro2024sm}. The primary clinical task is often to classify the presence or absence of lesions or disease in the overall organ being imaged (e.g. brain or lungs). A complementary task is \emph{localization}, where the goal is to indicate which specific 2D slices of the 3D volume show evidence for the overall prediction.
Localization models predict a (probabilistic) binary label for each slice.
Models with high-quality localization allow clinicians to audit predictions, hopefully improving interpretability, trust, and accountability.
Yet in many applied tasks, especially with 3D medical scans, the expense of acquiring expert labels means that model development can only be ``weakly'' supervised: coarse scan-level binary labels are available for each 3D scan in the training set, but not fine-grained labels for individual slices. An unsolved technical challenge is thus to develop effective slice-level localization from weak supervision, while preserving competitive whole-scan classification.

Deep multiple instance learning (MIL) is a strategy for such weakly-supervised learning~\citep{quellec2017multiple,saab2019doubly}. MIL methods can take as input a variable-sized set of instances (a ``bag''), each with their own feature vector, and produce one predicted binary label. For localization with deep MIL, attention-based MIL~\citep{ilse2018attention} has become the foundation of many methods in the last decade due to its by-design interpretability: each instance $j$ in bag $i$ has an attention weight $a_{ij} \in [0, 1]$ indicating its relative importance for the classification task. Overall, localization in MIL remains underexplored~\citep{castro2024sm}.
Past works have used attention weights for localization primarily through \emph{qualitative} results~\citep{ilse2018attention,lu2021data,shao2021transmil,keshvarikhojasteh2024multi} that visually highlight regions of interest (ROIs) with high attention values.
A few works also quantitatively evaluate  localization~\citep{li2021dual,fourkioti2024camil,castro2024sm}.


Recently, \citet{harvey2026multi} applied MIL to several datasets of 3D brain scans, including computed tomography (CT) and magnetic resonance imaging (MRI). \citeauthor{harvey2026multi}~showed a simple center-focused baseline that \emph{ignores all image content} can outperform the learned attention from modern MIL for localization.
The attention values for 3D scan $i$ in this baseline are set as $a_{ij} \propto \operation{NormPDF}(j | \frac{S_i}{2}, 1)$ where $S_i$ is the number of 2D axial slices in the 3D scan (``bag'') and integer $j$ indexes each 2D slice (``instance'') from 1 to $S_i$ in spatial order. This discretized univariate Normal PDF results in bell-shaped attention when plotted as a function of $j$, with a peak at the middle slice. It
reflects the clinical intuition that the center of the brain is more relevant than the edges for prediction tasks like detecting the presence of lesions or white matter disease. 

In this paper, we offer three contributions to 
improve the development of weakly-supervised classifiers of 3D medical images, focusing on CT scans due to large, high-quality available datasets:
\begin{itemize}[leftmargin=*,noitemsep,topsep=0pt]
    \item We show that the simple center-focused baseline from \citet{harvey2026multi} outperforms modern transformer-based MIL at slice localization tasks that span several body parts, including thoracic CT (lungs) and abdominal CT (kidneys, spleen, liver and bowel). This result expands the brain-specific results of past work and suggests current MIL methods are not adequate for localization in 3D medical imaging even given large labeled data. \emph{Better inductive biases} are clearly needed.

    \item Motivated by this baseline, we propose Normal Guidance for the attention mechanism within deep MIL. We use a normal distribution parameterized by the empirical mean and variance of the learned attention weights to regularize the learned attention weights. This inductive bias encourages the attention as a function of $j$ to form a smooth bell-curve shape. Normal Guidance does not force all scans to focus on the exact center; instead image-specific content can inform attention when helpful.
    We demonstrate that Normal Guidance enables attention-based and transformer-based MIL to perform better than the simple baseline for localization (Tab.~\ref{tab:instance-level_auroc}) while remaining competitive at whole scan classification (Tab.~\ref{tab:bag-level_auroc}).
    \item We develop methods for quantifying a best-in-class ceiling for localization (Sec.~\ref{sec:best_possible_instance-level_performance}) and scan-level classification (Sec.~\ref{sec:best_possible_bag-level_performance}). These are approximate practical upper bounds for what our chosen MIL architectures (pooled frozen embeddings with linear classifier heads) are capable of given the available training data. These methods unfairly using \emph{instance-level} labels to establish best-case performance for weakly-supervised MIL.
    While other MIL methods remain far from the localization ceiling on all tasks, we show that MIL with Normal Guidance comes much closer (within 0.03 AUROC of ceiling) on two CT tasks.
\end{itemize}
Our supporting evidence for these contributions is drawn from experiments on three open-access medical imaging datasets of CT scans totaling over 4 million 2D slices. The datasets span different body parts (head, chest, abdomen), and the classification tasks cover acute injuries, embolisms, and lesions.
Each dataset includes human expert annotations for every slice. All non-ceiling models are trained strictly in the weak-supervised regime, using only scan-level labels for training, early stopping, and hyperparameter selection; slice-level labels are only used to evaluate localization quality. 

We hope this work improves weakly-supervised localization for 3D imaging tasks and advances understanding of what inductive biases are needed for deep MIL to move from the lab to the clinic.
\section{Background and Related Works}
\label{sec:related_works}


MIL~\citep{dietterich1997solving,maron1997framework,quellec2017multiple} is a branch of weakly supervised learning.
In MIL, the training dataset $\mathcal{D} = \{(x_i, y_i)\}_{i=1}^N$ consists of $N$ bags, where each bag $x_i = \{x_{i,1}, \dots, x_{i,S_i}\}$ contains a set of $S_i$ instances.
Although instance-level labels exist $\{y_{i,1}, \dots, y_{i,S_i}\}$ where $\forall j : y_{ij} \in \{0, 1\}$, only the bag-level label $y_i \in \{0, 1\}$ is observed during training.
The standard MIL assumption~\citep{dietterich1997solving,raff2023reproducibility} is that a bag is negative if and only if \emph{all} instances are negative $\left( y_i = 0 \Leftrightarrow \forall j \in \{1, \dots, S_i\} : y_{ij} = 0 \right)$ and positive if and only if \emph{at least one} instance is positive $\left( y_i = 1 \Leftrightarrow \exists j \in \{1, \dots, S_i\} : y_{ij} = 1 \right)$.

MIL architectures typically consist of three parts: an encoder, pooling operation, and classifier.
First, an encoder $f(\cdot) : \mathbb{R}^{C \times H \times W} \rightarrow \mathbb{R}^M$ maps each $C$-channel 2D image $x_{ij} \in \mathbb{R}^{C \times H \times W}$ to an instance-level embedding $h_{ij} = f(x_{ij}) \in \mathbb{R}^M$.
Second, a pooling operation $\sigma(\cdot) : \mathbb{R}^{M \times S_i} \rightarrow \mathbb{R}^M$ maps the set of instance-level embeddings $h_i = \{h_{i,1}, \dots, h_{i,S_i}\}$ to a single bag-level embedding $z_i = \sigma(h_i) \in \mathbb{R}^M$.
Finally, a classifier $g(\cdot) : \mathbb{R}^M \rightarrow [0, 1]$ maps the bag-level embedding to a predicted probability $\hat{y}_i = g(z_i) \in [0, 1]$.

\textbf{Permutation-invariant pooling.}
Early MIL methods~\citep{pinheiro2015image,zhu2017deep,feng2017deep} used simple, non-learnable pooling operations to aggregate instance-level embeddings.
\citet{ilse2018attention} proposed learnable attention-based MIL (ABMIL) to be both data-driven and interpretable.
Extensions to ABMIL have maintained permutation-invariance (e.g., CLAM~\citep{lu2021data}, DSMIL~\citep{li2021dual}, MAD-MIL~\citep{keshvarikhojasteh2024multi}).

\textbf{Correlated MIL.} 
Permutation-invariance is inappropriate in cases with clear spatial dependencies across instances, such as bags of patches drawn from the same high-resolution 2D image or bags of 2D slices from a 3D volume. 
To address this, recent work has proposed \emph{correlated MIL}~\citep{shao2021transmil} where transformer-based MIL methods (e.g., TransMIL~\citep{shao2021transmil}, CAMIL~\citep{fourkioti2024camil}) use multi-head self-attention to incorporate dependencies between instances.

\textbf{Smooth operator.} 
To improve localization while accounting for dependencies, \citet{castro2024sm} proposed a smoothing operator that smoothes the learned embeddings of neighboring instances.
Their module achieved state-of-the-art localization results among all tested MIL methods on 3D neuroimage and whole slide imaging (WSI) classification tasks.

\textbf{Centered Gaussian.}
\citet{harvey2026multi} raised questions about the quality of modern MIL for localization in brain scans.
They examined a simple baseline: a discretized Gaussian distribution centered on the middle 2D slice of a 3D brain scan, $a_{ij} \propto \operation{NormPDF}(j | \frac{S_i}{2}, 1)$.
They found this baseline outperforms the smoothing operator and other attention-based and transformer-based MIL at localizing in 3D neuroimages.

\textbf{Alternatives for attention regularization.}
In MIL for WSI, prior works have proposed regularization techniques for learned attention weights.
\citet{sharma2021cluster} proposed cluster-to-cluster, which adds a Kullback-Leibler (KL) divergence term between the attention weights of patches in a cluster and the uniform distribution.
\citet{zhang2025aem} introduced attention entropy maximization, which subtracts an entropy term on the attention weights to penalize excessive attention concentration.
\citet{peled2026psa} integrates 2D spatial context into pairwise multi-head attention through a learnable distance-decayed prior. 
Unlike these methods, our Normal Guidance encourages a univariate learned attention distribution to follow a bell-shaped curve. 

For further expanded discussion of related work, see App.~\ref{app:related_work}.

\section{Methods}

\subsection{Guided Attention}

We formulate a guided attention framework designed to regularize the learned attention weights $a_i$ by penalizing their divergence from a discrete reference distribution $r_i$ over the $S_i$ instances in bag $i$.
The guided objective that we minimize during training is defined as:
\begin{align}
    \label{eq:guided_objective}
    \textstyle
    \mathcal{L}_{\text{Guided}} := \frac{1}{N} \sum_{i=1}^N \ell^{\text{BCE}}(y_i, \hat{y}_i ) + \lambda D(r_i, a_i)
\end{align}
where $\ell^{\text{BCE}}$ is the standard binary cross entropy loss, $\lambda \geq 0$ is a hyperparameter that controls the strength of the regularization term, and $D(r_i, a_i)$ is a divergence measure quantifying the ``distance'' between the reference distribution and the learned attention weights for bag $i$.
To construct the reference $r_i$, we recommend a Normal Guidance procedure described in Sec.~\ref{sec:normal_guidance}. We apply $D(r_i, a_i)$ to all bags (positive and negative), with each $r_i$ constructed from the attention values of bag $i$. 
Designing an $r_i$ that depends on bag label is possible, but we leave this to future work.
Some tasks could imagine using instance-level labels to inform the reference $r_i$ (see App.~\ref{app:label_guidance}), but this requires available labels and breaks the assumptions of MIL.

\begin{figure}[!t]
    \includegraphics[width=\linewidth]{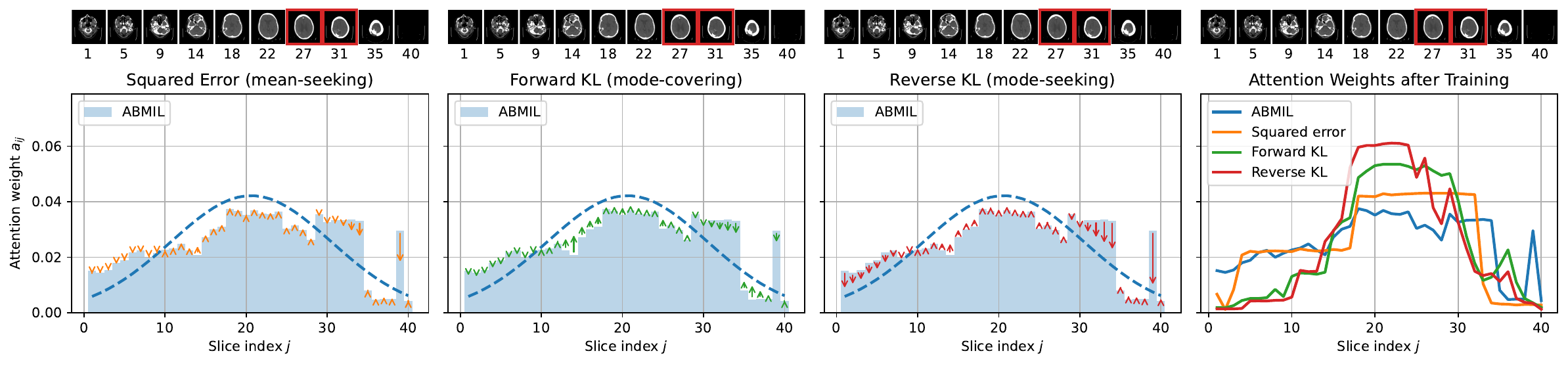}
    \caption{Normal Guidance on one 3D CT scan under different divergences $D$. \emph{Left panels}: using squared error ({\color{matplotlib_orange} orange}), forward KL ({\color{matplotlib_green} green}), and reverse KL ({\color{matplotlib_red} red}). Arrows indicate direction of change for each attention weight $a_{ij}$ to reduce $D(\cdot)$ via a gradient step; line length indicates magnitude of the regularization term for each attention weight. \emph{Right}: Attention weights $a_{i}$ after training with standard ABMIL or with Normal Guidance. Normal Guidance improves the distribution's unimodality and focus, though recovering the true ROI (red region in top bar) remains imperfect.}
    \label{fig:varying_the_divergence_measure}
\end{figure}

Among possible choices for divergence $D(\cdot)$, we study three variants with different behavior, described below and compared side-by-side in Fig.~\ref{fig:varying_the_divergence_measure}:
\begin{itemize}[leftmargin=*,noitemsep,topsep=0pt]
    \item \textbf{Squared error (mean-seeking).} The squared error $D(r_i, a_i) = \sum_{j=1}^{S_i} (r_{ij} - a_{ij})^2$ is symmetric, penalizes large errors more than small errors, and generally encourages  $a_i$ to be \emph{mean-seeking}.
    \item \textbf{Forward KL (mode-covering).} The forward KL $D_{\text{KL}}(r_i \| a_i) = \sum_{j=1}^{S_i} r_{ij} \log \frac{r_{ij}}{a_{ij}}$ is zero-avoiding with respect to $r_i$, resulting in \emph{mode-covering} behavior.
    As illustrated in Fig.~\ref{fig:varying_the_divergence_measure}, it asymmetrically penalizes the model most if $a_{ij}$ is small when $r_{ij}$ is large.
    See \citet{murphy2012forward} for more details.
    \item \textbf{Reverse KL (mode-seeking).} The reverse KL $D_{\text{KL}}(a_i \| r_i) = \sum_{j=1}^{S_i} a_{ij} \log \frac{a_{ij}}{r_{ij}}$ is zero-forcing with respect to $a_i$, resulting in \emph{mode-seeking} behavior.
    As illustrated in Fig.~\ref{fig:varying_the_divergence_measure}, it asymmetrically penalizes the model most if $a_{ij}$ is large when $r_{ij}$ is small.
    See \citet{murphy2012forward} for more details.
\end{itemize}

\subsection{Normal Guidance}
\label{sec:normal_guidance}

In many \emph{correlated MIL} tasks, instances are spatially or temporally dependent.
For example, in medical imaging, if a tumor is visible in one slice, it is likely to be present in adjacent slices.
To incorporate this inductive bias into the guided attention framework, we propose Normal Guidance.
First, we compute the empirical mean and variance of the attended slice position under the learned distribution $a_i$:
\begin{align}
    \textstyle \mathbb{E}[J] = \sum_{j=1}^{S_i} j \cdot a_{ij}, \qquad \textstyle \operation{Var}(J) = \left( \sum_{j=1}^{S_i} j^2 \cdot a_{ij} \right) - \left( \sum_{j=1}^{S_i} j \cdot a_{ij} \right)^2.
\end{align}
Next, these values define a Normal PDF that we evaluate at integers $j$ from 1 to $S_i$ and renormalize to obtain the reference distribution $r_i$:
\begin{align}
\label{eq:normal_reference}
    \hat{r}_{ij} = \operation{NormPDF}(j | \mathbb{E}[J], \operation{Var}(J)), \qquad r_{ij} = \frac{\hat{r}_{ij}}{\sum_{k=1}^{S_i} \hat{r}_{ik}}.
\end{align}
When performing gradient descent on the loss in Eq.~\eqref{eq:guided_objective}, we use the stop-gradient operation on $r_i$
to require updates to $a_i$ that improve agreement with $r_i$. We recompute $r_i$ before each backward pass.

\subsection{Multi-Head Normal Guidance}

\begin{figure}[!b]
    \centering
    \includegraphics[width=0.5\linewidth]{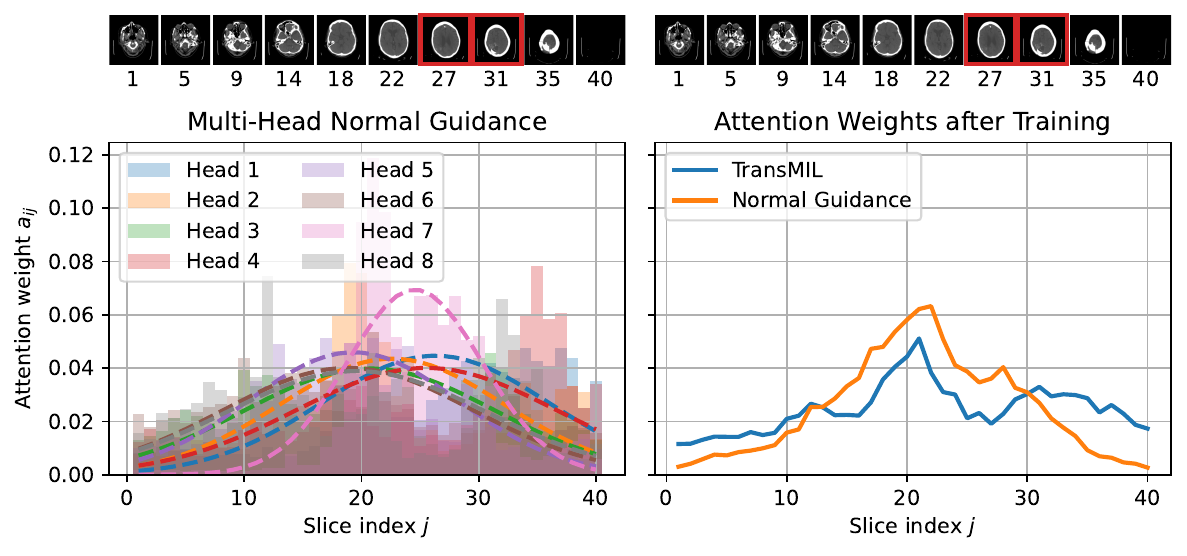}
    \caption{Multi-Head Normal Guidance. Left: discretized normal distribution for each attention head in TransMIL (we set $H=8$ throughout). Right: Mean attention weight $a_{ij}$ for each head after training, for standard TransMIL and TransMIL plus Multi-Head Normal Guidance.
    }
    \label{fig:multi-head_normal_guidance}
\end{figure}

While Normal Guidance incorporates local dependencies, it encourages learning \emph{unimodal} attention weights.
This can be restrictive in correlated MIL settings where multiple disjoint regions are relevant (e.g., multiple tumors across non-adjacent slices).
In such cases, a single normal distribution may over-smooth across regions or focus on only one region.

To overcome this unimodal restriction, we extend our method to Multi-Head Normal Guidance. 
For multi-head self-attention, we use a separate unimodal reference distribution via Eq.~\eqref{eq:normal_reference} for each head of the final self-attention layer (see Fig.~\ref{fig:multi-head_normal_guidance}).
The overall regularization term is computed by averaging the divergence across all heads.
Because each head is guided toward a separate bell-shaped curve, the combined effect can be intuitively thought of as guided by a Gaussian mixture model.
This extension allows the network to simultaneously attend to multiple spatially separate regions. 

\section{Experiments}
\label{sec:experiments}

\begin{figure}[!t]
    \centering
    \includegraphics[width=1\linewidth]{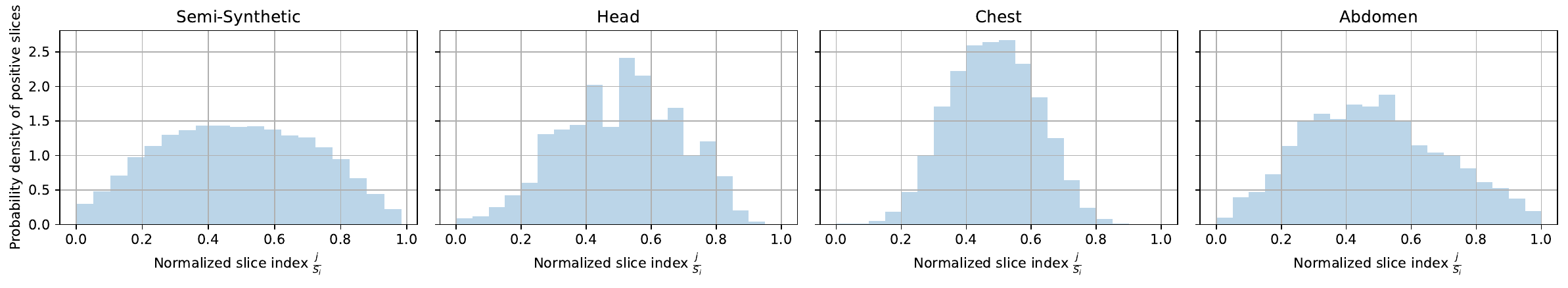}
    \caption{Probability density of positive slices in training set. Because slices in the middle of the scan are more likely to be positive the centered Gaussian performs well for localization.}
    \label{fig:probability_density_of_positive_slices}
\end{figure}

\subsection{Datasets}
\label{sec:datasets}

\textbf{Head CT.}
The RSNA 2019 Brain CT Hemorrhage Challenge~\citep{flanders2020construction} includes 752,803 slices from 21,744 CT scans (20-60 slices per scan) with labels for presence/absence of any intracranial hemorrhage (ICH).
We partition the released training set (no other release has labels) into our own training, validation, and test sets as described below.
\citet{castro2024sm} train and test on a smaller subset of this data (1,150 scans total).

\textbf{Chest CT.}
The RSNA Pulmonary Embolism CT dataset~\citep{colak2021rsna} has 1,790,594 slices from 7,279 CT scans (63-1,083 slices per scan) with labels for presence/absence of any pulmonary embolism.
We partition the released training set (no other release has labels) as described below.

\textbf{Abdomen CT.}
The RSNA 2023 Abdominal Trauma AI Challenge~\citep{hermans2024rsna} includes 1,500,653 slices from 4,711 CT scans (41-1,727 slices per scan) with labels for presence/absence of any abdominal trauma (AT).
We partition the released training set (no other release has labels) as described below.

\textbf{Partition by patient.}
For each medical dataset (Head CT, Chest CT, and Abdomen CT), we randomly assign 3D scans into training, validation, and testing sets using a 4:1:1 ratio, ensuring each patient’s data belongs to exactly one set to avoid leakage.
We stratify by class to ensure comparable class frequencies.
We repeat this process across three random seeds, so every learning algorithm is evaluated on the same 3 distinct partitions into training, validation, and test sets.

\textbf{Expert annotations.} All 3 CT scan datasets include a binary label for each whole CT scan \emph{and} an instance-level binary label for each 2D axial slice in every scan. The Head CT dataset was annotated by 60 neuroradiologists; the Chest CT dataset was annotated by a group of more than 80 expert thoracic radiologists; and the Abdomen CT dataset was annotated by radiologists with subspecialty training and/or professional experience in abdominal trauma. In our work, slice-level labels are only used to evaluate localization and establish best-in-class performance ceilings.

\textbf{Semi-synthetic.} The \emph{Shifted Mean MIL} dataset~\citep{harvey2025synthetic,harvey2026multi} was created to assess generalization capabilities of MIL methods.
We follow prior work's generative process: we draw $y_i \sim \operation{Bern}(0.5)$ so 50\% of bags are positive and $S_i \sim \operation{Unif}(\{20, \dots, 60\})$ so bags have between 20 and 60 instances.
If a bag is negative, all $M$ features of all $S_i$ instances are drawn from a mean 0, variance 1 Gaussian.
If a bag is positive, we select a contiguous block of $R = 12$ instances and draw only the first of $M$ features in this block from a Gaussian with mean $\Delta = 0.5$; all other features use a mean 0, variance 1 Gaussian. 
We call this \emph{semi-}synthetic because the range of $S_i$ and the value of $R$ are set to match the Head CT dataset, and the number of features $M=768$ is set to match the ViT embedding size we use. 
We sample 10,000 bags for training, 2,500 bags for validation, and 1,000 bags for testing.
We repeat this process with three random seeds; each seed samples a different training, validation, and test set. Further details are in App.~\ref{app:semisynth}.

\subsection{Experimental Goals, Metrics, Baselines, and Training Details}
\label{sec:training_details}

We are primarily interested in assessing the \emph{localization performance} of MIL methods trained with weak supervision. Following past work~\citep{castro2024sm}, we measure this performance only on positive-labeled bags in the test set. For each bag $i$ such that $y_i = 1$, at all $S_i$ slices indexed by $j$ we assess the predicted attention value $a_{ij}$ versus the ground truth label $y_{ij}$ via the AUROC, using macro averaging over bags. While our focus is localization, we also wish to verify that scan-level classification remains strong. We thus report scan-level (``bag''-level) classification quality in terms of AUROC. For both localization and bag-level classification, AUPRC results are in 
App.~\ref{app:auprc_results}.
The primary reported number (AUROC or AUPRC) is the \emph{average} over the 3 test partitions of each task. The number after the ``$\pm$`` symbol indicates the empirical standard deviation of the 3 values.

\textbf{Baselines.}
Among MIL methods, we consider max pooling MIL, mean pooling MIL, attention-based MIL (ABMIL;~\citealp{ilse2018attention}), transformer-based MIL (TransMIL;~\citealp{shao2021transmil}), and the Smooth Operator (known as SmAP and SmTP from \citet{castro2024sm}). 
We chose not to compare to CAMIL~\citep{lu2021data}, DSMIL~\citep{li2021dual}, and CLAM~\citep{fourkioti2024camil} because of their poor localization results on the Head CT dataset reported in \citeauthor{castro2024sm}, where these methods were outperformed by the Smooth Operator in all tested localization tasks.

For localization only, we also compare to the center-focused Gaussian baseline from \citeauthor{harvey2026multi}.
This method predicts a centered bell-shaped curve across slices for every scan.
It is not informed at all by image content and cannot produce scan-level predictions.
See App.~\ref{app:localization} for details about localization experiments.

For all experiments with Normal Guidance, for simplicity we set $D$ to the forward KL and fix regularization strength $\lambda=1$; we conduct a sensitivity analysis of these choices later in Fig.~\ref{fig:sensitivity_auroc}.
We configure baseline methods by following past work, providing reproducible details in App.~\ref{app:modeldetails}.

\textbf{Encoder.}
For all medical imaging experiments, we use a \emph{frozen} ViT-B/16~\citep{dosovitskiy2021image} pre-trained on ImageNet~\citep{deng2009imagenet} as the raw feature encoder, with embedding size $M=768$.
Past work~\citep{harvey2026multi} found MIL with a frozen ViT encoder to be competitive with other encoders like ConvNeXt or MedSAM.
Their frozen ViT with MIL pipeline was also competitive with fully fine-tuned 3D CNNs while being far more affordable to train. For all methods and tasks, we feed a final scan-level embedding to a linear binary classifier head.

\textbf{Training.}
All methods are optimized using mini-batch stochastic gradient descent (SGD) with a momentum parameter of 0.9 and batch size of 64.
We train for 1,000 epochs and use bag-level validation AUROC for early stopping.
For all methods, we grid search two key hyperparameters: learning rate in \{0.1, 0.01, 0.001, 0.0001\} and L1 regularization strength for all pooling and classifier weights in \{1.0, 0.1, 0.01, 0.001, 0.0001, 1e-5, 1e-6, 0.0\}.

\subsection{Best-in-Class Ceiling for Instance-Level Localization}
\label{sec:best_possible_instance-level_performance}

On any benchmark, it is important to understand how close the current methods are to ``saturation'', meaning the best possible performance on that task by a model in that class given the designated inputs. The key assumptions of our approach are using a linear classification head on frozen ViT embeddings. Here we try to establish a ``best-in-class ceiling'' for each task that represents a realistic upper bound for any correlated MIL method using our assumptions. 
This is not merely a classifier that handles each instance separately with a ``one feature in, one prediction out'' architecture, used as an upper bound in some past MIL works~\citep{correia2020automatic,guillaumin2010multiple}. Instead, our instance-level ceiling for CT consumes a local neighborhood of embeddings to account for spatial dependencies in the correlated MIL setting.

\textbf{Ceiling for CT tasks.} We train an instance-level classifier using ground truth instance-level labels $y_{ij}$. This classifier takes as input for instance index $j$ a local window of $R$ consecutive embedding vectors centered on $h_{ij}$ and outputs an instance-level predicted probability $\hat{y}_{ij}$. We use a 1D convolutional layer with kernel size $R$ over instance representations provided by the frozen ViT, followed by a linear classifier head.
For each dataset, we set $R$ equal to the mean number of consecutive positive instances, allowing gaps of up to 3 negative instances in a ``consecutive'' block.
This sets $R=12$ for Head CT, $R=35$ for Chest CT, and $R=29$ for Abdomen, according to Tab.~\ref{tab:describe_blocks}.

\textbf{Ceiling for semi-synthetic.}
For \emph{Shifted Mean MIL} data, we derive the probability $p(y_{ij}{=}1 | h_i, S_i, y_i{=}1)$: the chance that instance $j$ comes from the Gaussian with \emph{shifted mean} given its features and the fact that its bag is positive. This event $y_{ij}{=}1$ is equivalent to index $j$ falling in the chosen latent segment of $R$ consecutive instances. The generative model is framed in terms of the starting index $u_i$ of this segment, so let $\mathcal{U}_{ij}$
 be the set of possible values of $u_i$ that could make $j$ positive.
 Adding up the probability of these relevant values, we can write 
 $p(y_{ij}{=}1 | h_i, S_i, y_i{=}1)$ as
\begin{align}
    \label{eq:marginal}
    = \sum_{u_i \in \mathcal{U}_{ij}} p(u_i | h_i, S_i, y_i{=}1) = \sum_{u_i \in \mathcal{U}_{ij}} \frac{p(h_i | u_i, S_i, y_i{=}1)p(u_i | S_i, y_i{=}1)}{p(h_i | S_i, y_i{=}1)}.
\end{align}
This uses Bayes' rule. Each term in Eq~\eqref{eq:marginal} can be computed in closed-form, as explained in App.~\ref{sec:bayes_estimator_for_localization}.
This oracle upper bound for instance-level performance has not been derived in prior work.

\subsection{Best-in-Class Ceiling for Bag-Level Classification}
\label{sec:best_possible_bag-level_performance}

\textbf{Bag ceiling for CT tasks.}
We establish an upper bound for bag-level performance of MIL by using oracle instance-level labels $y_{ij}$ to pool embeddings, instead of learned attention. This of course has an unfair advantage compared to weakly-supervised MIL since it relies on instance-level annotations \emph{even at test time} and not just images alone. However, our intention is to establish a ceiling for image-only methods, since we hope the ideal attention matches expert labels.

Procedurally, 
for each positive bag, we pool instance-level embeddings using a uniform distribution over that bag's positively labeled instances.
For each negative bag, we randomly sample a positive bag and use its instance-level label distribution, adjusting for differing slice counts via interpolation.
We then pool the negative bag's embeddings uniformly over all ``positive'' instances. 
This enforces similar ROI distributions regardless of class to create scan-level representations, which are used to train a scan-level linear binary classifier.
As an alternative to enforcing similar ROIs, we tried averaging over all negative bag slices uniformly but found unintended shortcuts were possible due position-specific feature distributions.
If negative bags pool over the entire volume but positive bags concentrate on specific regions, a simple ``how dark is this image?'' feature could detect the more dark slices at index 1 or $S_i$ in negative bags and yield near-perfect classification that is too optimistic.

\textbf{Bag ceiling for semi-synthetic.}
Since the true data-generating process is known, we use the Bayes estimator derived in Eq.~(10) of \citet{harvey2026multi} as an upper bound on bag-level performance.

\section{Results}
\label{sec:results}

\subsection{Localization performance}
\label{sec:experiments_loc}

\begin{table}[!b]
    \caption{Localization AUROC on test set, aka slice-level or instance-level classification, using positive bags only. Takeaway: On real CT images, Normal Guidance outperforms all other MIL and is only MIL to significantly beat the image-free Centered Gaussian baseline.}
    \label{tab:instance-level_auroc}
    \centering
    \resizebox{\linewidth}{!}{\begin{tabular}{lcccc}
        \hline
        \bfseries Method & \bfseries Semi-Synthetic & \bfseries Head CT & \bfseries Chest CT & \bfseries Abdomen CT \\
        \hline
        Centered Gaussian & $0.689${\tiny$\pm 0.015$} & $0.850${\tiny$\pm 0.001$} & $0.780${\tiny$\pm 0.002$} & $0.573${\tiny$\pm 0.031$} \\
        \hline
        \hline
        Max pooling & $0.504${\tiny$\pm 0.002$} & $0.286${\tiny$\pm 0.004$} & $0.461${\tiny$\pm 0.003$} & $0.453${\tiny$\pm 0.011$} \\
        Mean pooling & $0.500${\tiny$\pm 0.000$} & $0.500${\tiny$\pm 0.000$} & $0.500${\tiny$\pm 0.000$} & $0.500${\tiny$\pm 0.000$} \\
        \hline
        ABMIL & $0.546${\tiny$\pm 0.054$} & $0.736${\tiny$\pm 0.052$} & $0.558${\tiny$\pm 0.086$} & $0.526${\tiny$\pm 0.041$} \\
        ~~Smooth Operator & $0.693${\tiny$\pm 0.030$} & $0.831${\tiny$\pm 0.003$} & $0.688${\tiny$\pm 0.043$} & $0.564${\tiny$\pm 0.012$} \\
        ~~Normal Guidance (ours) & $0.541${\tiny$\pm 0.055$} & \boldsymbol{$0.871${\tiny$\pm 0.002$}} & $0.866${\tiny$\pm 0.005$} & \boldsymbol{$0.663${\tiny$\pm 0.006$}} \\
        \hline
        TransMIL & $0.593${\tiny$\pm 0.131$} & $0.792${\tiny$\pm 0.031$} & $0.585${\tiny$\pm 0.077$} & $0.579${\tiny$\pm 0.003$} \\
        ~~Smooth Operator & $0.592${\tiny$\pm 0.127$} & $0.823${\tiny$\pm 0.019$} & $0.654${\tiny$\pm 0.035$} & $0.582${\tiny$\pm 0.043$} \\
        ~~Multi-Head Normal Guidance (ours) & \boldsymbol{$0.706${\tiny$\pm 0.076$}} & $0.869${\tiny$\pm 0.002$} & \boldsymbol{$0.869${\tiny$\pm 0.002$}} & $0.634${\tiny$\pm 0.007$} \\
        \hline
        \hline
        Best-in-Class Ceiling (see Sec.~\ref{sec:best_possible_instance-level_performance}) & $0.884${\tiny$\pm 0.004$} & $0.896${\tiny$\pm 0.003$} & $0.871${\tiny$\pm 0.004$} & $0.717${\tiny$\pm 0.011$} \\ 
        \hline
    \end{tabular}}
\end{table}

Localization results are in Tab.~\ref{tab:instance-level_auroc} and \ref{tab:instance-level_auprc}. 
The major findings from these results are listed below.

\textbf{The Centered Gaussian baseline outperforms existing MIL for localization.}
Tab.~\ref{tab:instance-level_auroc} and \ref{tab:instance-level_auprc} show that the simple baseline from \citet{harvey2026multi} not only outperforms the attention mechanism for localization on the Head CT task but also on the thoracic and abdominal tasks. On Chest CT, no existing MIL scores above 0.69 while the Centered baseline scores 0.78!

\textbf{Normal Guidance can outperform this baseline: a naive center bias is not always best.}
The centered Gaussian $a_{ij} \propto \operation{NormPDF}(j | \frac{S_i}{2}, 1)$ encodes the inductive bias that positive instances are most likely in the center of the bag
(see probability density of positive slices in Fig.~\ref{fig:probability_density_of_positive_slices}).
Yet our Normal Guidance (NG) and Multi-Head Normal Guidance (MHNG) 
deliver notable gains in AUROC over this baseline of at least +0.019 on Head CT, +0.08 on Chest CT, and +0.06 on Abdomen CT. This reassuringly indicates that learning attention from image features can outperform the image-agnostic Centered Gaussian baseline.


\subsection{Bag-level classification performance}

Bag-level (scan-level) classification is reported in Tab.~\ref{tab:bag-level_auroc} and ~\ref{tab:bag-level_auprc}. 
Recall that Normal Guidance is specifically designed to improve localization. Here our goal is to verify how this  regularization impacts bag-level classification and interpret how methods compare to our new ceilings.

\textbf{Normal Guidance keeps scan-level classification competitive.} Across all tested CT tasks, our NG and MHNG score better than or within tolerance (within 0.01 AUROC) of their corresponding unregularized or smoothed alternative. On semi-synthetic bag-level classification, ABMIL + Smooth Operator scores highest (0.771) but TransMIL with MHNG is competitive at 0.766.

\textbf{On two CT tasks, the best MIL methods are near the bag-level ceiling.}
The best MIL method achieves very good results on Head CT (within 0.002 AUROC of best-in-class ceiling) and good results on Chest CT (within 0.022 AUROC of the best-in-class ceiling).
For the semi-synthetic dataset, the best MIL method is within 0.042 AUROC of its ceiling.
For the Abdomen CT task, the gap of about 0.1 AUROC may be real. Yet we guess the ceiling is overestimated; this task has much more variability in the organs covered than other tasks, and 6.6\% of scans do not fully cover one or more organs.
Using instance-level labels from an incomplete positive scan to pool a complete negative scan (or vice versa) could create shortcut features that overestimate performance.

\textbf{Transformer-based MIL only slightly improves bag-level performance.} 
Transformer-based methods improve AUROC by up to 0.05 on the Abdomen CT task over ABMIL-based counterparts, but otherwise don't seem to add appreciable gain on the other 3 tasks.

\begin{table}[!h]
    \caption{Bag-level AUROC on test set for whole-scan classification.
    Takeaway: Normal Guidance is competitive with state-of-the-art MIL alternatives here, while improving localization (Tab.~\ref{tab:instance-level_auroc}).
    }
    \label{tab:bag-level_auroc}
    \centering
    \resizebox{\linewidth}{!}{\begin{tabular}{lcccc}
        \hline
        \bfseries Method & \bfseries Semi-Synthetic & \bfseries Head CT & \bfseries Chest CT & \bfseries Abdomen CT \\
        \hline
        Max pooling & $0.624${\tiny$\pm 0.009$} & $0.888${\tiny$\pm 0.009$} & $0.656${\tiny$\pm 0.004$} & $0.643${\tiny$\pm 0.029$} \\
        Mean pooling & $0.752${\tiny$\pm 0.002$} & $0.920${\tiny$\pm 0.012$} & $0.669${\tiny$\pm 0.018$} & $0.625${\tiny$\pm 0.027$} \\
        \hline
        ABMIL & $0.751${\tiny$\pm 0.003$} & $0.919${\tiny$\pm 0.009$} & $0.664${\tiny$\pm 0.017$} & $0.639${\tiny$\pm 0.044$} \\
        ~~Smooth Operator & $0.771${\tiny$\pm 0.008$} & $0.925${\tiny$\pm 0.012$} & $0.671${\tiny$\pm 0.023$} & $0.648${\tiny$\pm 0.043$} \\
        ~~Normal Guidance (ours) & $0.751${\tiny$\pm 0.003$} & $0.925${\tiny$\pm 0.010$} & $0.678${\tiny$\pm 0.026$} & $0.651${\tiny$\pm 0.026$} \\
        \hline
        TransMIL & $0.763${\tiny$\pm 0.016$} & $0.925${\tiny$\pm 0.014$} & $0.663${\tiny$\pm 0.023$} & $0.684${\tiny$\pm 0.047$} \\
        ~~Smooth Operator & $0.768${\tiny$\pm 0.025$} & $0.926${\tiny$\pm 0.013$} & $0.670${\tiny$\pm 0.015$} & $0.664${\tiny$\pm 0.050$} \\
        ~~Multi-Head Normal Guidance (ours) & $0.766${\tiny$\pm 0.006$} & $0.926${\tiny$\pm 0.012$} & $0.662${\tiny$\pm 0.015$} & $0.677${\tiny$\pm 0.029$} \\
        \hline
        \hline
        Best-in-Class Ceiling (see Sec.~\ref{sec:best_possible_bag-level_performance}) & $0.810${\tiny$\pm 0.006$} & $0.927${\tiny$\pm 0.012$} & $0.700${\tiny$\pm 0.014$} & $0.776${\tiny$\pm 0.033$} \\
        \hline
    \end{tabular}}
\end{table}

\subsection{Sensitivity Analysis}

\begin{figure}[!t]
    \centering
    \noindent
    \begin{minipage}[c]{0.6\linewidth}
        \centering
        \begin{tabular}{lcc}
            \hline
            \bfseries Divergence & \bfseries Localization & \bfseries Classification \\
            \hline
            Squared error 
            & $0.834${\tiny$\pm 0.007$} & $0.924${\tiny$\pm 0.012$} \\
            Forward KL 
            & $0.871${\tiny$\pm 0.002$} & $0.925${\tiny$\pm 0.010$} \\
            Reverse KL 
            & $0.869${\tiny$\pm 0.003$} & $0.919${\tiny$\pm 0.011$} \\
            \hline
        \end{tabular}
    \end{minipage}%
    \begin{minipage}[c]{0.35\linewidth}
        \centering
        \includegraphics[width=\linewidth]{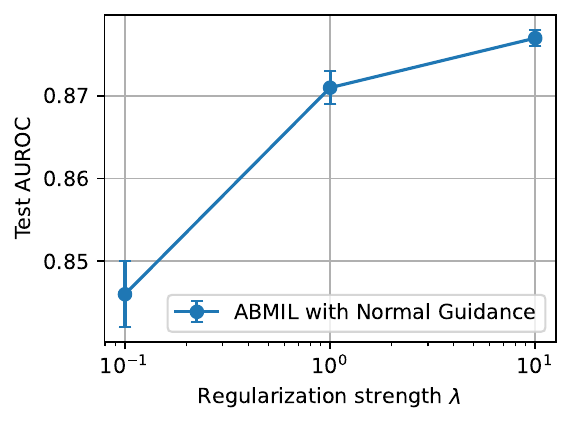}
    \end{minipage}
    \caption{Sensitivity analysis. \emph{Left:} Effect of divergence $D$ on localization and bag-level classification, with $\lambda=10^0$.
    \emph{Right:} Effect of regularization strength $\lambda$ on localization ($D$: Forward KL). 
    All results here report test set AUROC on the Head CT task, using ABMIL with Normal Guidance.
    }
    \label{fig:sensitivity_auroc}
\end{figure}

Our Normal Guidance procedure's success could depend on the choice of divergence $D$ and the regularization strength $\lambda$. Thus far, these values have been fixed as forward KL and $\lambda = 1$.
The two panels of Fig.~\ref{fig:sensitivity_auroc} show how performance on the Head CT task depends on these variables.
First, the left panel of Fig.~\ref{fig:sensitivity_auroc} suggests a modest preference for Forward KL for both localization and whole-scan classification.
The right panel of Fig.~\ref{fig:sensitivity_auroc} suggests that $\lambda=10^0 = 1$ is a reasonable choice for localization, but we might do slightly better by tuning $\lambda$. Thus, the localization success of NG and MHNG reported in Sec.~\ref{sec:experiments_loc} might be improvable.

\subsection{Qualitative Side-by-Side Comparisons of Learned Attention}

See App.~\ref{app:attention_viz} for visual comparisons of the learned attention distributions from different MIL methods on a handful of randomly-selected CT scans from each of our 3 datasets. In each figure, we either compare ABMIL, ABMIL with Sm, and ABMIL with NG; or TransMIL, TransMIL with Sm, and TransMIL with MHNG.
As a primary takeaway, we can visually confirm that Normal Guidance often produces more \emph{focused} and unimodal attention distributions than its counterparts.
Some plots, especially the right panels of Fig.~\ref{fig:abdomen_ct_attention_visualizations}, show cases where all methods fail to capture the key block of positive slices, indicating plenty of room for further localization method development.

\section{Conclusion}
\label{sec:conclusion}

The attention mechanism was originally proposed in MIL as a method that incorporates interpretability~\citep{ilse2018attention}. Yet this work shows that for localization in 3D medical images, MIL alone cannot outperform a simple baseline due to a lack of inductive bias.
We proposed Normal Guidance, a regularization technique for attention-based MIL models that encourages attention to be bell-curve shaped.
We investigated three CT datasets totaling over 4 million 2D slices and show that Normal Guidance outperforms all other attention-based MIL methods as well as the simple slice-level baselines, achieving a new state-of-the-art for instance-level MIL results on all datasets while maintaining high bag-level performance.

\textbf{Limitations.}
(1)~We only use positive bags to evaluate slice-level localization, following past work~\citep{castro2024sm}.
We do not evaluate the quality of attention in negative bags.
This highlights an open problem in the literature: defining what attention should look like for a negative bag.
(2) \citet{jain2019attention} argue that attention weights should not be treated as faithful explanations of classifier decisions, since permuting attention can yield similar predictions.
Our results show that Normal Guidance can help learned attention track expert annotations more closely, but this does not by itself enforce that the attended slices are causally responsible for the bag-level prediction. (3) We focus on deep MIL with frozen encoders and linear classifiers, again following past work~\citep{lu2021data,shao2021transmil,fourkioti2024camil,castro2024sm}. Parameter-efficient fine-tuning could deliver more effective encoders for this domain, but at prohibitive computational cost. Even with frozen encoders, model training and hyperparameter optimization takes roughly 9 hours for each TransMIL run on  the head CT dataset (not to mention repeating this across 32 hyperparameter configurations and 3 train/test partitions).

\textbf{Future work.}
Beyond 3D images, MIL is often used for whole slide images. Alternative inductive biases that account for the 2D spatial dependencies in WSI are critical for understanding  localization results in that domain. A second direction is conditioning $r_i$ on the bag label $y_i$: the right reference (Gaussian, Uniform, etc.) for negative bags is worth investigating.  Finally, the same bell-shaped prior (NG) may also suit MIL tasks beyond medical imaging, such as video anomaly detection, where evidence is typically contiguous in time \citep{sultani2018real}.

\section*{Acknowledgments}
This work is supported by the U.S. National Institutes of Health (grant \# R01NS134859) and the Alzheimer’s Drug Discovery Foundation. Author MCH is also supported in part by the U.S. National Science Foundation (NSF) via IIS CAREER grant \# 2338962. 
We are grateful for resources and support from the Tufts High-Performance Computing Cluster.
This paper's content is solely the responsibility of the authors and does not necessarily represent the official views of the NIH or NSF.

\bibliographystyle{plainnat}
\bibliography{main}

\appendix
\counterwithin{table}{section}
\setcounter{table}{0}
\counterwithin{figure}{section}
\setcounter{figure}{0}

\newpage 
\section{Dataset Details}

\subsection{Preprocessing of CT scans}

For all medical imaging datasets (Head CT, Chest CT, and Abdomen CT), we convert images into Hounsfield Units (HU) using each image's rescale slope and intercept; resize each 2D slice to $224 \times 224$ pixels; and normalize images with the training set mean and standard deviation of each channel.
For the Head CT dataset, we use intensity windowing to exclude the skull, other bones, and calcifications (only including -100 to 300 HU) \citep{muschelli2019recommendations}.

\subsection{Descriptive Statistics}

Below, we include three tables that summarize the CT datasets:
\begin{itemize}[leftmargin=*,noitemsep,topsep=0pt]
    \item the number of  blocks (ROIs) in a bag,
    \item the number of instances in a typical block (ROI), and
    \item the fraction of all instances in the bag contained in a typical block (ROI).
\end{itemize}

\begin{table}[htbp!]
    \centering
    \begin{tabular}{lccccccc}
        \hline
        & \bfseries 1 block & \bfseries 2 blocks & \bfseries 3 blocks & \bfseries 4 blocks & \bfseries 5 blocks & \bfseries 6 blocks & \bfseries 7 blocks \\
        \hline
        Head CT & 8296/8882 & 569/8882 & 17/8882 & -- & -- & -- & -- \\
        Chest CT & 1437/2211 & 497/2211 & 185/2211 & 65/2211 & 16/2211 & 7/2211 & 4/2211 \\
        Abdomen CT & 304/365 & 53/365 & 7/365 & 1/365 & -- & -- & -- \\
        \hline
    \end{tabular}
    \caption{Number of  blocks of consecutive positive instances in a bag, allowing gaps of up to 3 negative instances in a ``consecutive'' block.}
\end{table}
\begin{table}[htbp!]
    \centering
    \begin{tabular}{lcccccccc}
        \hline
        & \bfseries Mean & \bfseries Min & \bfseries 5th & \bfseries 25th & \bfseries 50th & \bfseries 75th & \bfseries 95th & \bfseries Max \\
        \hline
        Head CT & 12 & 1 & 2 & 6 & 11 & 17 & 23 & 43 \\
        Chest CT & 35 & 1 & 1 & 7 & 16 & 44 & 137 & 353 \\
        Abdomen CT & 29 & 1 & 2 & 6 & 15 & 40 & 95 & 242 \\
        \hline
    \end{tabular}
    \caption{Number of positive instances in a block. We allow gaps of up to 3 negative instances in a ``consecutive'' block.}
    \label{tab:describe_blocks}
\end{table}
\begin{table}[htbp!]
    \centering
    \begin{tabular}{lcccccccc}
        \hline
        & \bfseries Mean & \bfseries Min & \bfseries 5th & \bfseries 25th & \bfseries 50th & \bfseries 75th & \bfseries 95th & \bfseries Max \\
        \hline
        Head CT & 0.350 & 0.017 & 0.053 & 0.172 & 0.342 & 0.513 & 0.688 & 1.000 \\
        Chest CT & 0.093 & 0.001 & 0.004 & 0.023 & 0.059 & 0.137 & 0.283 & 0.499 \\
        Abdomen CT & 0.119 & 0.002 & 0.006 & 0.027 & 0.063 & 0.174 & 0.401 & 0.870 \\
        \hline
    \end{tabular}
    \caption{Fraction of all instances in a block (region of interest or ROI). We allow gaps of up to 3 negative instances in a ``consecutive'' block.}
\end{table}

\newpage 
\section{AUPRC Results}
\label{app:auprc_results}

\begin{table}[h!]
    \caption{Localization results for the \textbf{AUPRC} metric, using the test set of each task. Corresponding AUROC results are in Tab.~\ref{tab:instance-level_auroc}.}
    \label{tab:instance-level_auprc}
    \centering
    \resizebox{\linewidth}{!}{\begin{tabular}{lcccc}
        \hline
        \bfseries Method & \bfseries Semi-Synthetic & \bfseries Head CT & \bfseries Chest CT & \bfseries Abdomen CT \\
        \hline
        Centered Gaussian & $0.569${\tiny$\pm 0.012$} & $0.710${\tiny$\pm 0.007$} & $0.481${\tiny$\pm 0.008$} & $0.222${\tiny$\pm 0.038$} \\
        \hline
        \hline
        Max pooling & $0.384${\tiny$\pm 0.003$} & $0.302${\tiny$\pm 0.007$} & $0.185${\tiny$\pm 0.002$} & $0.109${\tiny$\pm 0.002$} \\
        Mean pooling & $0.332${\tiny$\pm 0.003$} & $0.359${\tiny$\pm 0.009$} & $0.188${\tiny$\pm 0.001$} & $0.104${\tiny$\pm 0.007$} \\
        \hline
        ABMIL & $0.424${\tiny$\pm 0.045$} & $0.634${\tiny$\pm 0.049$} & $0.233${\tiny$\pm 0.059$} & $0.159${\tiny$\pm 0.027$} \\
        ~~Smooth Operator & $0.574${\tiny$\pm 0.035$} & $0.713${\tiny$\pm 0.007$} & $0.344${\tiny$\pm 0.037$} & $0.167${\tiny$\pm 0.020$} \\
        ~~Normal Guidance (ours) & $0.403${\tiny$\pm 0.062$} & \boldsymbol{$0.744${\tiny$\pm 0.010$}} & $0.522${\tiny$\pm 0.008$} & \boldsymbol{$0.248${\tiny$\pm 0.020$}} \\
        \hline
        TransMIL & $0.487${\tiny$\pm 0.140$} & $0.659${\tiny$\pm 0.032$} & $0.258${\tiny$\pm 0.052$} & $0.191${\tiny$\pm 0.021$} \\
        ~~Smooth Operator & $0.467${\tiny$\pm 0.147$} & $0.689${\tiny$\pm 0.007$} & $0.309${\tiny$\pm 0.018$} & $0.195${\tiny$\pm 0.039$} \\
        ~~Multi-Head Normal Guidance (ours) & \boldsymbol{$0.578${\tiny$\pm 0.086$}} & $0.740${\tiny$\pm 0.010$} & \boldsymbol{$0.535${\tiny$\pm 0.005$}} & $0.227${\tiny$\pm 0.010$} \\
        \hline
        \hline
        Best-in-Class Ceiling (see Sec.~\ref{sec:best_possible_instance-level_performance}) & $0.825${\tiny$\pm 0.005$} & $0.781${\tiny$\pm 0.010$} & $0.528${\tiny$\pm 0.004$}  & $0.285${\tiny$\pm 0.030$} \\ 
        \hline
    \end{tabular}}
\end{table}

\begin{table}[!h]
    \caption{Bag-level \textbf{AUPRC} on test set for whole-scan classification.
    Takeaway: Normal Guidance is competitive with state-of-the-art MIL alternatives here, while improving localization (Tab.~\ref{tab:instance-level_auprc}).
    Corresponding AUROC results are in Tab.~\ref{tab:bag-level_auroc}.}
    \label{tab:bag-level_auprc}
    \centering
    \resizebox{\linewidth}{!}{\begin{tabular}{lcccc}
        \hline
        \bfseries Method & \bfseries Semi-Synthetic & \bfseries Head CT & \bfseries Chest CT & \bfseries Abdomen CT \\
        \hline
        Max pooling & $0.616${\tiny$\pm 0.005$} & $0.859${\tiny$\pm 0.013$} & $0.440${\tiny$\pm 0.002$} & $0.133${\tiny$\pm 0.016$} \\
        Mean pooling & $0.756${\tiny$\pm 0.007$} & $0.903${\tiny$\pm 0.011$} & $0.455${\tiny$\pm 0.005$} & $0.117${\tiny$\pm 0.010$} \\
        \hline
        ABMIL & $0.755${\tiny$\pm 0.009$} & $0.903${\tiny$\pm 0.007$} & $0.451${\tiny$\pm 0.005$} & $0.122${\tiny$\pm 0.014$} \\
        ~~Smooth Operator & $0.775${\tiny$\pm 0.010$} & $0.910${\tiny$\pm 0.011$} & $0.465${\tiny$\pm 0.008$} & $0.139${\tiny$\pm 0.014$} \\
        ~~Normal Guidance (ours) & $0.755${\tiny$\pm 0.010$} & $0.911${\tiny$\pm 0.008$} & $0.477${\tiny$\pm 0.015$} & $0.117${\tiny$\pm 0.008$} \\
        \hline
        TransMIL & $0.766${\tiny$\pm 0.019$} & $0.912${\tiny$\pm 0.012$} & $0.447${\tiny$\pm 0.013$} & $0.139${\tiny$\pm 0.025$} \\
        ~~Smooth Operator & $0.770${\tiny$\pm 0.025$} & $0.913${\tiny$\pm 0.012$} & $0.461${\tiny$\pm 0.009$} & $0.135${\tiny$\pm 0.007$} \\
        ~~Multi-Head Normal Guidance (ours) & $0.766${\tiny$\pm 0.015$} & $0.912${\tiny$\pm 0.012$} & $0.451${\tiny$\pm 0.010$} & $0.139${\tiny$\pm 0.012$} \\
        \hline
        \hline
        Best-in-Class Ceiling (see Sec.~\ref{sec:best_possible_bag-level_performance}) & $0.814${\tiny$\pm 0.002$} & $0.913${\tiny$\pm 0.010$} & $0.468${\tiny$\pm 0.004$} & $0.252${\tiny$\pm 0.081$} \\
        \hline
    \end{tabular}}
\end{table}

\begin{figure}[!h]
    \centering
    \noindent
    \begin{minipage}[c]{0.65\linewidth}
        \centering
        \begin{tabular}{lcc}
            \hline
            \bfseries Divergence & \bfseries Localization & \bfseries Classification \\
            \hline
            Squared error 
            & $0.702${\tiny$\pm 0.011$} & $0.910${\tiny$\pm 0.010$} \\
            Forward KL 
            & $0.744${\tiny$\pm 0.010$} & $0.911${\tiny$\pm 0.008$} \\
            Reverse KL 
            & $0.747${\tiny$\pm 0.012$} & $0.905${\tiny$\pm 0.011$} \\
            \hline
        \end{tabular}
    \end{minipage}%
    \begin{minipage}[c]{0.35\linewidth}
        \centering
        \includegraphics[width=\linewidth]{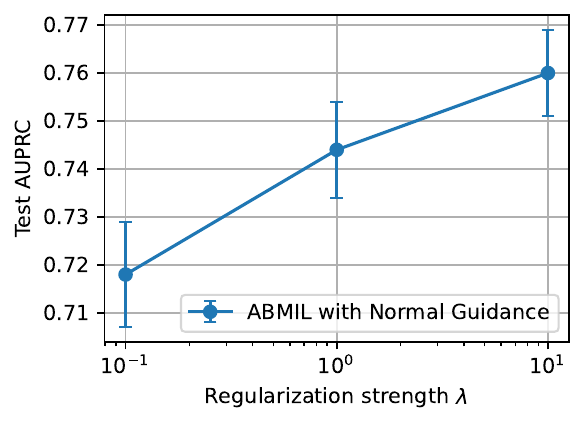}
    \end{minipage}
    \caption{Sensitivity analysis for the \textbf{AUPRC} metric on the test set.
     \emph{Left:} Effect of divergence $D$ on localization and bag-level classification, with $\lambda=10^0$.
    \emph{Right:} Effect of regularization strength $\lambda$ on localization ($D$: Forward KL). 
    All results from the Head CT task, using ABMIL with Normal Guidance.
    }
    \label{fig:sensitivity_auprc}
\end{figure}

\newpage
\section{Attention Visualizations}
\label{app:attention_viz}

Below, you can see visuals of the learned attention weights for key MIL methods on the Head, Chest, and Abdomen CT tasks.

\begin{figure}[!h]
    \centering
    \includegraphics[width=\linewidth]{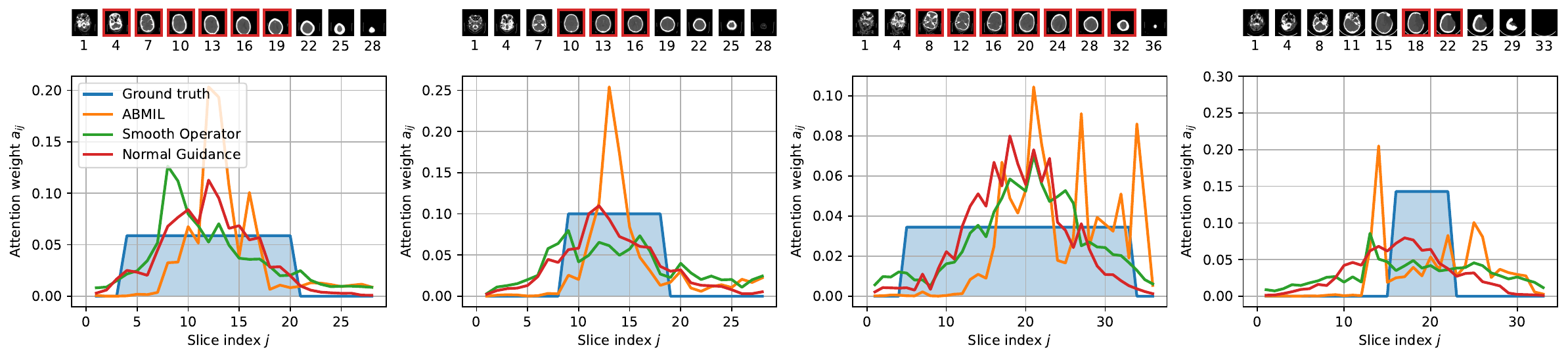}
    \includegraphics[width=\linewidth]{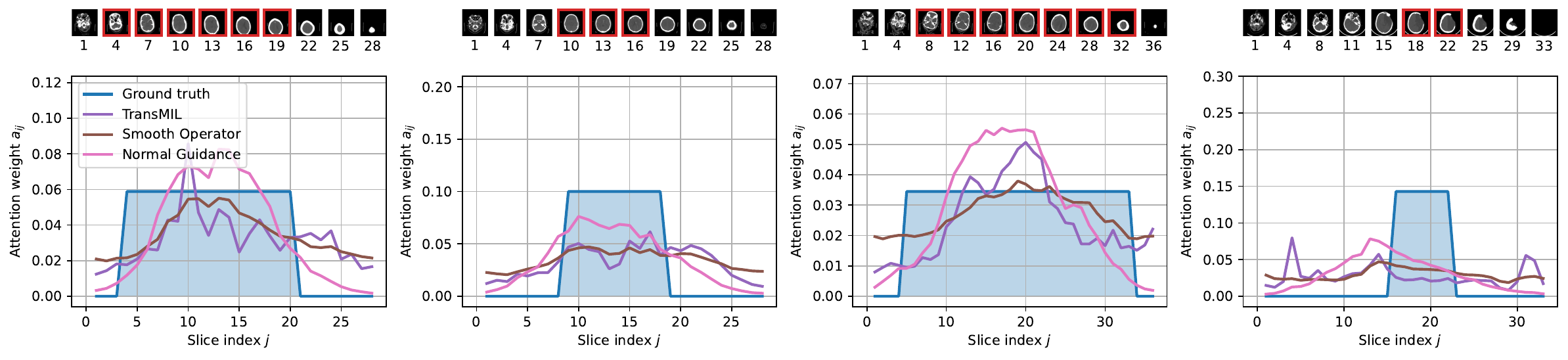}
    \caption{Learned attention weights for 4 example CT scans (\emph{columns}) from the \textbf{Head} CT dataset. \emph{Top row}: methods based on ABMIL. \emph{Bottom row}: transformer-based MIL methods.
    }
    \label{fig:head_ct_attention_visualizations}
\end{figure}

\begin{figure}[!ht]
    \centering
    \includegraphics[width=\linewidth]{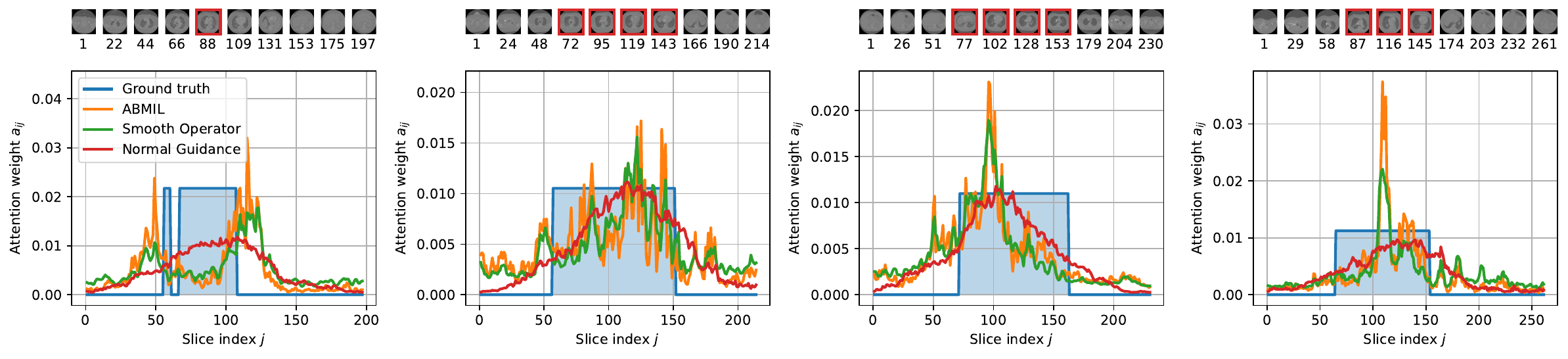}
    \includegraphics[width=\linewidth]{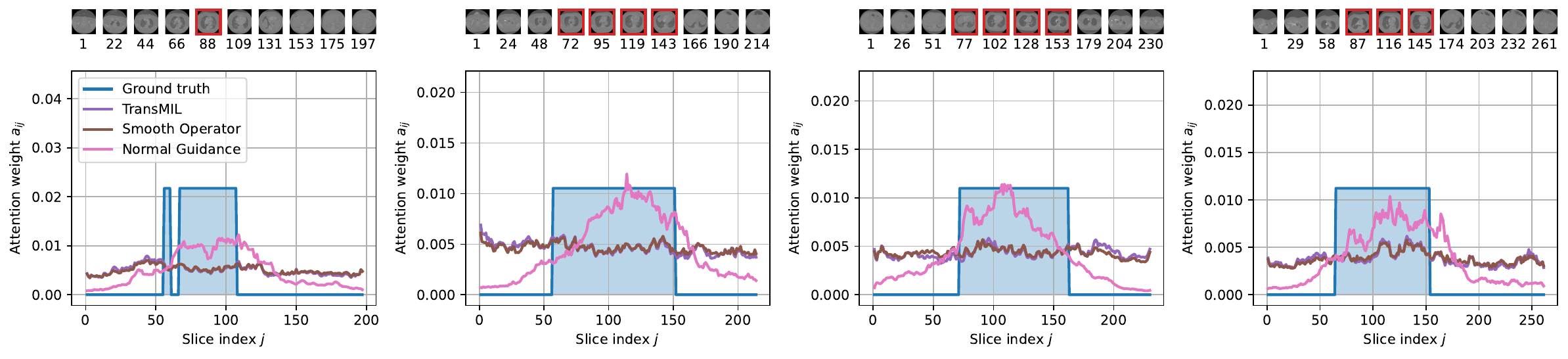}
    \caption{Learned attention weights for 4 example CT scans (\emph{columns}) from the \textbf{Chest} CT dataset. \emph{Top row}: methods based on ABMIL. \emph{Bottom row}: transformer-based MIL methods.
    }
    \label{fig:chest_ct_attention_visualizations}
\end{figure}

\begin{figure}[!ht]
    \centering
    \includegraphics[width=\linewidth]{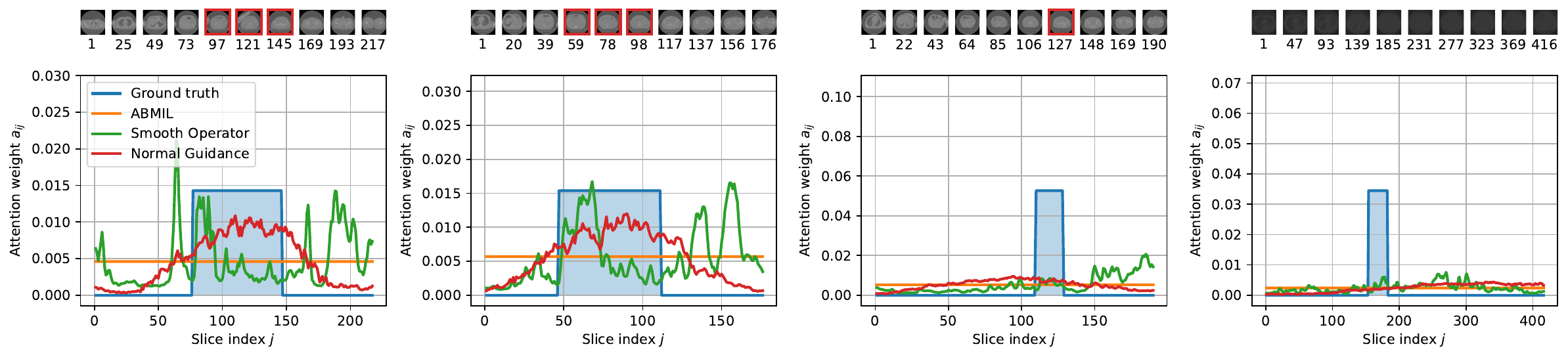}
    \includegraphics[width=\linewidth]{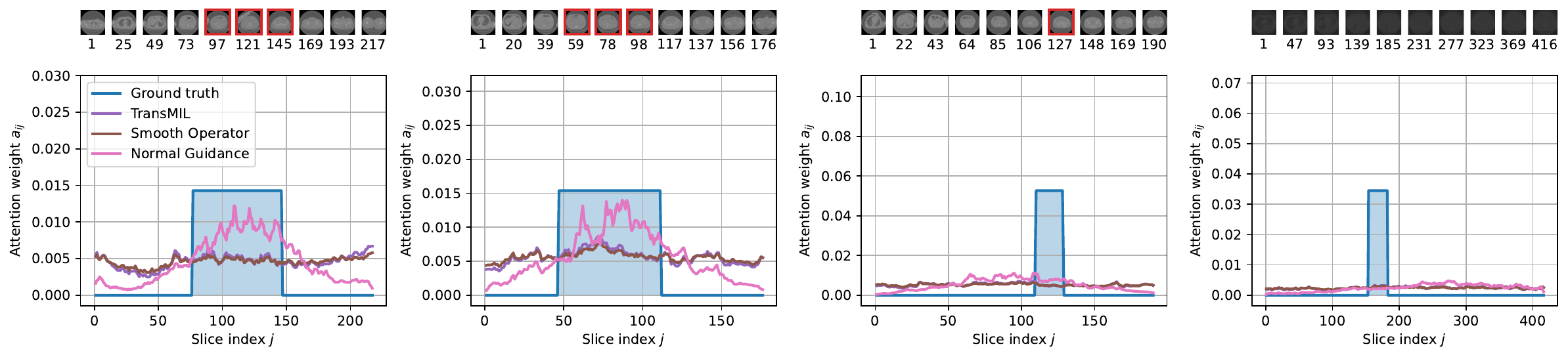}
    \caption{Learned attention weights for 4 example CT scans (\emph{columns}) from the \textbf{Abdomen} CT dataset. \emph{Top row}: methods based on ABMIL. \emph{Bottom row}: transformer-based MIL methods.
    }
    \label{fig:abdomen_ct_attention_visualizations}
\end{figure}

\newpage

\section{Further Details about Baseline Models and Experiments}
\label{app:modeldetails}

\subsection{Compute Resources}
All experiments use either 4 Intel Xeon 6342 CPUs (2.80 GHz) or 4 Intel Xeon 6226R CPUs (2.90 GHz) and either 1 NVIDIA RTX 6000 (24 GB), 1 NVIDIA RTX A6000 (48 GB), or 1 NVIDIA A100 GPU (40 GB).

\subsection{Localization Experiment Details}
\label{app:localization}

For most MIL methods, the attention score $a_{ij}$ used to evaluate localization is straightforward.
For max pooling MIL only, we need to construct a post-hoc attention score for each instance. 
Because we use element-wise max pooling MIL, we can construct attention weights $\alpha_{ij} = \frac{1}{M} \sum_{k=1}^M \mathds{1}[\operation{argmax}_{j^\prime}(h_{ij^\prime k})=j]$ where $\mathds{1}[\cdot]$ denotes the indicator function.

\subsection{Model Details}

All MIL models use a frozen ViT-B/16 encoder ($M=768$); only the pooling head and
linear classifier are trained.

\paragraph{Smooth Operator (\textsc{Sm}).}
We use an approximation of \textsc{Sm} following~\citet{castro2024sm},
defined by $T=10$ iterations of
$g^{(t+1)} = (1-\alpha)\,h_i + \alpha\,A\,g^{(t)}$ starting from
$g^{(0)}=h_i$. Here $A$ is the
normalized adjacency matrix of the chain graph representing the
slice ordering of the 3D scan. The mixing scalar is parameterized as
$\alpha=\sigma(\tilde\alpha)$ with $\tilde\alpha\in\mathbb{R}$ learnable and
initialized so $\alpha_{t=0}=0.5$.  For SmAP, a single \textsc{Sm} module is
applied to the encoder's output embeddings prior to ABMIL.
For SmTP, a separate \textsc{Sm} module (each with its own learnable
$\alpha$) is inserted after each MHSA layer.

\paragraph{TransMIL.}
We follow~\citet{shao2021transmil}: a learnable class token is prepended to
$h_i$, followed by two multi-head self-attention blocks separated by a Pyramidal Position Encoding Generator (PPEG)
module. The class-token output of the second block is the bag representation.
Unlike the reference implementation, which approximates self-attention with
Nystr\"om attention, we use full scaled dot-product self-attention
(\texttt{torch.nn.MultiheadAttention}). Both blocks use $H=8$ heads at
model dim $M=768$ (per-head dim $96$) with no dropout. PPEG is three
parallel depthwise 1D convolutions with kernel sizes $\{3,5,7\}$, stride $1$,
and \textsc{same} padding, summed and added to the non-class-token positions.
Reported slice-level attention $a_{i,j}$ is the class-token row of the second
block's attention map, averaged over heads, with the class-token-to-class-token
entry removed.  
\section{Semi-Synthetic Data and its Bayes Estimator for Localization}
\label{app:semisynth}
\label{sec:bayes_estimator_for_localization}

For our data-generating process, we set $M=768$ and $R=12$. 
We sample $S_i$ uniformly between 20 and 60 for any bag regardless of its label.
We draw $y_i$, the bag-level binary label, from a Bernoulli distribution with probability 0.5.

For negative bags, we draw all $S_i$ feature vectors $h_{ij} \in \mathbb{R}^M$ from the same simple distribution: each entry is drawn iid from a Gaussian with mean 0 and variance 1.

For positive bags, we draw the starting index $u_i$ of the modified $R$ adjacent instances from a uniform distribution over the valid range
\begin{align}
    p(u_i | S_i, y_i{=}1) = \operation{UnifPMF}(u_i | \{1, \dots, S_i{-}R{+}1\}).
\end{align}
Next, given the $u_i$ for a positive bag we draw only the first feature (of $M$) of every instance in the chosen block of size $R$ from a Gaussian with \emph{shifted-mean} $\Delta = 0.5$. Otherwise, all features are drawn from the same zero-mean, unit variance Gaussian that defines the negative bags:
\begin{align}
    p(h_i | u_i, S_i, y_i{=}1) = \prod_{j=1}^{S_i} \prod_{k=1}^{M}
    \begin{cases}
        \operation{NormPDF}(h_{ijk} | 0.5, 1), 
        & \text{if $j \in [u_i, u_i{+}R{-}1]$ and $k = 1$}
        \\
        \operation{NormPDF}(h_{ijk}|0, 1), & \text{otherwise.}
    \end{cases}
\end{align}

Using the above as ingredients, we can compute the marginal likelihood of an observed feature vector alone via the sum rule, summing over all possible starting indices
\begin{align}
    p(h_i | S_i, y_i{=}1) = \sum_{u=1}^{S_i{-}R{+}1} p(h_i | u_i = u, S_i, y_i{=}1) p(u_i = u | S_i, y_i{=}1).
\end{align}
The three terms above allow computation of the Bayes estimator for the instance-level label via the right-hand-side of Eq.~\eqref{eq:marginal}.

\textbf{Relevant set of $u_i$ values for index $j$.}
In the main paper, we define the set $\mathcal{U}_{ij}$ as the set of possible $u_i$ values that would make index $j$ fall in the chosen latent segment of $R$ consecutive instances. We can define this set more formally as
$$
\mathcal{U}_{ij} = \{u_i \in \mathbb{Z} : \max(1, j - R + 1) \leq u_i \leq \min(S_i - R + 1, j)\} 
$$
We know the size of the set is always at most R: $|\mathcal{U}_{ij}| \leq R$. It will only be smaller than $R$ for indices $j$ within $R$ steps of either edge.

\section{Label Guidance}
\label{app:label_guidance}

An assumption in MIL is that instance-level labels are unknown during training.
However, if instance-level labels $\{y_{i,1}, \dots, y_{i,S_i}\}$ are known, they can be used to define a discrete reference distribution $r_i$ to regularize the learned attention weights $a_i$.
For example, for a negative bag we could define a uniform distribution over all instances $p(j | y_i{=}0) = \operation{UnifPMF}(j|\{1, \dots, S_i\})$ and for a positive bag we could define a uniform distribution over all positive instances $p(j | y_i{=}1) = \operation{UnifPMF}(j|\{k \in \{ 1, \dots, S_i \} : y_{i,k}{=}1\})$.
These attention weights would achieve a perfect score for instance-level AUROC and AUPRC.
Recent work has used instance-level semantic labels to guide attention~\citep{liu2024semantics}.
Other works have used instance-level semantic labels to train classifiers to supervise attention~\citep{huang2023detecting}.
\section{Additional Related Work}
\label{app:related_work}

\textbf{Inductive biases in MIL.}
Other work in MIL has pursued inductive bias in the dependencies across instances. To account for 2D spatial relations among instances in MIL for whole slide imaging (WSI), \citet{peled2026psa}'s probabilistic spatial attention (PSA) method modifies self-attention with a learnable distance-decayed prior. This makes nearby instances (patches) likely to attend to each other in the $S_i \times S_i$ pairwise self-attention matrix. Unlike our approach with a regularization term that is agnostic to any pooling strategy, their approach directly modifies the construction of attention values and relies on multi-head self-attention as a specific attention mechanism. PSA also does not necessarily encourage unimodality, which seems to be a beneficial inductive bias in 3D CT datasets we tackle here.

\textbf{Gaussian and GMM ideas in MIL.}
Distantly related work \citep{zhang2021modeling} outside of MIL has used Gaussian mixture ideas to determine how one token should pay attention to other tokens in a text sequence for NLP translation tasks, hoping to encourage concentrated attention. Our work avoids the expense of a distinct GMM for every instance in the sequence, instead forming just one Gaussian-like discrete distribution per head. Our work is also distinct in its purpose to improve localization for MIL in 3D CT settings.

Our work is somewhat related to other efforts in MIL to incorporate probabilistic modeling ideas, such as work using Gaussian Processes to predict instance-level and bag-level labels given permutation-invariant per-instance embeddings \citep{wu2021combining}. That work uses Gaussian machinery to model a latent response function which is then transformed into a binary label; they do not regularize attention as we do toward helpful inductive biases such as unimodality and smoothness.

\subsection{Discussion of Past work on Upper Bounds using Instance-Level Labels}

\citet{correia2020automatic} measure the performance of a permutation-invariant MIL on in-the-wild speech in Youtube videos to detect Parkinson's or depression.
They develop a ``fully supervised'' upper bound for MIL via a classifier that consumes one instance at a time, somehow comparing this classifier given bags of size 1 to MIL classifiers given bags of larger size in the same figure. \citet{guillaumin2010multiple} consider a similar instance-level upper bound in a metric-based MIL approach for multi-label MIL problems. 
Instead, our work develops separate best-in-class ceilings for both instance-level and bag-level tasks, uses a local neighborhood for the 1D CNN instance-level classifier instead of a ``one instance in and one label out'' design, and always evaluates competitor methods on bags of the same size (much larger than one).

\subsection{Alternatives to Weak Supervision}

An alternative to weakly-supervised MIL with zero annotation costs is fully-unsupervised anomaly detection. Unsupervised methods for detecting brain lesions from 3D MRI scans have been proposed using diffusion models ~\citep{behrendt2024patched} or self-supervised learning where the pretext task is predicting the location of a given patch within the 2D axial slice~\citep{baker2025patch2loc}. These approaches generally rely on an expensive bespoke-trained representation rather than frozen encoders that are more easily ported across medical image modalities, and also are likely less accurate at scan-level prediction than the deep MIL studied here due to the lack of supervised labels informing model training.





\end{document}